\documentclass[a4paper]{article}

\usepackage{INTERSPEECH2021}
\usepackage{graphicx}
\usepackage{url}
\usepackage{hyperref}

\title{Language Proficiency and F0 Entrainment: A Study of L2 English Imitation in Italian, French, and Slovak Speakers}
\name{Zheng Yuan$^{1,2}$, 
{\v{S}}tefan Be{\v{n}}u{\v{s} $^{3,4}$}, 
Alessandro D'Ausilio$^{1,2}$}
\address{
  $^1$Italian Institute of Technology, Italy\\
  $^2$University of Ferrara, Italy\\
  $^3$Constantine the Philosopher University in Nitra, Slovakia\\ 
  $^4$Institute of Informatics, SAS, Slovakia}
\email{
\{zheng.yuan, alessandro.dausilio\}@iit.it,
sbenus@ukf.sk}

\begin{document}

\maketitle
\begin{abstract}
  This study explores F0 entrainment in second language (L2) English speech imitation during an Alternating Reading Task (ART). Participants with Italian, French, and Slovak native languages imitated English utterances, and their F0 entrainment was quantified using the Dynamic Time Warping (DTW) distance between the parameterized F0 contours of the imitated utterances and those of the model utterances. Results indicate a nuanced relationship between L2 English proficiency and entrainment: speakers with higher proficiency generally exhibit less entrainment in pitch variation and declination. However, within dyads, the more proficient speakers demonstrate a greater ability to mimic pitch range, leading to increased entrainment. This suggests that proficiency influences entrainment differently at individual and dyadic levels, highlighting the complex interplay between language skill and prosodic adaptation.
\end{abstract}

\section{Introduction}
\label{sec:intro}

Speech entrainment \cite{levitan2011measuring}, often termed alignment \cite{pickering2004toward}, convergence \cite{pardo2006phonetic}, or accommodation \cite{giles_accommodation_1991}, is observed as individuals unconsciously or deliberately modify their speaking style during interactions, leading to heightened similarity in acoustic-prosodic features. 

Entrainment, functioning as implicit imitation, is theorised to share neural-cognitive mechanisms with speech imitation \cite{gambi2013prediction}, involving automatic priming \cite{pickering2004toward}, motor control \cite{sato2013converging} and the perception-production loop \cite{goldinger1998echoes}. Dialect formation and language change stem from multiple imitative speech interactions \cite{delvaux2007influence}.

Studies have shown that entrainment plays a positive role in second language (L2) acquisition. Learners can effectively adapt to the target language by aligning their pronunciation with either native (L1) speakers \cite{jiang2022impact} or proficient L2 speakers \cite{trofimovich2014interactive}. Gnevsheva et al. \cite{gnevsheva2021phonetic} suggest second language learners exhibit greater flexibility in adapting their pronunciation compared to native speakers. Jiang et al. \cite{jiang2022impact} found that L2 learners' belief in an interlocutor's proficiency can influence L2 phonetic entrainment, ultimately bringing them closer to native speakers with an improved vowel pronunciation. Additionally, the entrainment effect varies based on individual differences, such as language talent \cite{lewandowski2019phonetic} and context (e.g., task type). Imitative tasks exhibit a higher degree of entrainment than interactive tasks \cite{dejong22_interspeech, yuan23b_interspeech}, particularly for non-native (L2) sounds \cite{wilt2023automatic}. 

Despite extensive research focusing on phonetic entrainment in L1-L2 interactions, the prosodic entrainment in L2-L2 scenarios remains largely unexplored, particularly considering the potential influence of L2 proficiency. Utilising the Alternating Reading Task (ART) corpus \cite{yuan2024art}, an L2 English speech dataset featuring recordings from Italian, French, and Slovak speakers, we conducted subjective evaluations to assess spoken English proficiency and subsequently delve into its connection with the fundamental frequency (F0) entrainment in L2-L2 speech imitation. Our investigation focuses on global static proximity and synchrony in entrainment as framed by \cite{levitan2011measuring, wynn2022classifying}. 

Past research on global static entrainment predominantly involves statistical analyses, including paired t-tests applied to "partner distance" and "other distance"\cite{levitan2011measuring}, as well as techniques like Time-aligned Moving Average (TAMA) \cite{kousidis2008towards}, Cross-Recurrence Quantification Analysis \cite{fusaroli2014analyzing}, and Windowed Lagged Cross-Correlation \cite{boker2002windowed}. While effective, these methods have limitations in controlling the unit of analysis, such as phonemes or words.

We employed a novel hybrid method that involves aligning timestamps at the word level using the WhisperX \cite{bain23_interspeech} ASR tool, adopting F0 contour parameterization \cite{reichel2018entrainment, reichel2023copasul}, and computing F0 entrainment as the distance between two parameterized F0 contours using Dynamic Time Warping (DTW). Three experiments were performed to validate the robustness of the F0 entrainment measurement algorithm and examine the correlation between F0 entrainment and L2 proficiency at both the individual speaker and dyadic levels.

\section{Data}
\subsection{The ART corpus}
The Alternating Reading Task (ART) corpus \cite{yuan2024art} is a collection of recordings from a collaborative L2 English speech production experiment designed to investigate entrainment and imitation. The corpus encompasses data gathered from 58 subjects, organised into same-sex dyads, with 18 native Italian speakers (6 males), 20 native French speakers (all female), and 20 native Slovak speakers (10 males). The experiment includes three distinct conditions: solo, interactive, and imitative utterance reading. Participants engaged in these tasks by reading utterances individually, taking turns in interactive sessions, and mimicking their dyadic partner's delivery of the target utterance during the imitative condition.

The textual material utilised in the experiment comprises a simplified adaptation of a Wikipedia article, totalling 801 words and segmented into 80 speaking turns. These speaking turns range from 6 to 13 words, and turn boundaries were strategically positioned within sentences to enhance prosodic continuity. During the experiment, participants were seated side by side, facing two screens, and separated by a curtain to mitigate the potential influence of mutual visual contact on speech entrainment. Notably, the focus of our analysis primarily revolves around the recordings from the imitation condition. 

\subsection{L2 proficiency evaluation}
To discern the potential influence of interlocutors' L2 speaking proficiency on speech entrainment, six language experts were enlisted to evaluate the spoken English skills of subjects in the ART corpus. The assessment focused on the initial 10 utterances from the solo recordings for each speaker, based on four key criteria: pronunciation, intonation, fluency, and overall impression. For each criterion, evaluators assigned scores on a scale ranging from 1 to 5, and a final score was derived as the average across the four criteria.

\begin{table}[th]
\caption{Intraclass Correlation Coefficients for Spoken English Proficiency Assessments}
\centering
\begin{tabular}{@{}llll@{}}
\toprule[1pt] 
\textbf{Indicator} & \textbf{ICC} & \textbf{p-value} & \textbf{CI95\%} \\ 
\midrule
pronunciation & 0.828 & $<0.001$ & [0.71, 0.90] \\
intonation    & 0.767 & $<0.001$ & [0.65, 0.85] \\
fluency       & 0.796 & $<0.001$ & [0.68, 0.87] \\
overall       & 0.800 & $<0.001$ & [0.67, 0.88] \\
final         & 0.840 & $<0.001$ & [0.73, 0.91] \\
\bottomrule[1pt] 
\end{tabular}%
\label{tab:icc}
\end{table}

Table \ref{tab:icc} delineates the degree of agreement among experts for each criterion, quantified through Intraclass Correlation Coefficients (ICC) values and their corresponding 95\% confidence intervals. The ICC values were computed using a two-way mixed-effects model with the mean of raters and 57 degrees of freedom. All ICC values indicate a level of "good reliability" (ICC between 0.75 and 0.9) with statistical significance (p-value $<$ 0.001), aligning with the criteria stipulated by \cite{koo2016guideline}. 


\section{Method}
\label{sec:method}
\subsection{Word segmentation and alignment}
Our approach to speech imitation analysis relied on utterance-level F0 contour comparison, comprising 4,640 (58 $\times$ 80) audio segments of speech imitation data. All instances of spoken words, inclusive of stutters, repetitions, and self-corrections, were retained.

To enhance the precision of time-series comparison, we executed word segmentation for each utterance. Audio files underwent transcription and force-alignment using the WhisperX ASR tool \cite{bain23_interspeech}, providing precise starting and ending timestamps for each word. The models employed for this task were the \texttt{base.en} Whisper model and the \texttt{Base\_960h} phoneme model.

The selected models exhibited a commendable precision score of 93.1 \cite{bain23_interspeech} on word segmentation for the Switchboard-1 Telephone Corpus (SWB), instilling confidence in their suitability for our analysis on the ART corpus. Through visual inspection by the authors, a comparable precision rate was observed for the ART corpus, with minor inaccuracies primarily associated with the cutting of the initial vowel in certain words. It's worth mentioning that manual calibration of the alignment was not conducted.

\subsection{F0 Preprocessing}
The extraction of F0 was performed utilising autocorrelation in PRAAT software through its Python interface Parselmouth (version 0.4.3) \cite{jadoul2018introducing}, with all parameters set to their default values. For precise and smooth F0 contours, voiceless utterance segments underwent linear interpolation, and a two-pass method \cite{de2009automatic} was implemented to address F0 outliers. Subsequently, a Savitzky-Golay filter was employed to enhance the smoothness of the F0 contour \cite{reichel2018entrainment, reichel2023copasul}, adopting third-order polynomials in 7-sample windows.

\subsection{Parameterization}
To explore the static global entrainment, specifically proximity and synchrony, as defined by \cite{levitan2011measuring, wynn2022classifying}, we adopted a simplified CoPaSul \cite{reichel2018entrainment, reichel2023copasul} style F0 parameterization. Five widely recognised and easily interpretable F0 features were selected for this purpose.

\begin{itemize}
    \item \textbf{Mean:} the average of pitch values $\mathbf{y}$.
    \item \textbf{Median:} the median of pitch values $\mathbf{y}$.
    \item \textbf{Slope:} the slope of the first-order linear fit of pitch values $\mathbf{y}$ against normalised time $\mathbf{t_{n}}$.
    \item \textbf{Range:} the difference between the 95th and 5th percentiles of the fitted values $\mathbf{y_{n}}$.
    \item \textbf{Drop:} the difference between the last and first fitted values $\mathbf{y_{n}}$, normalised by time $\mathbf{t}$.
\end{itemize}

These selected features serve distinct roles in the analysis: mean and median pertain to a more stable indicator of the physiological quality of speakers' articulators, while slope, range, and drop contribute to the dynamics of prosody. This parameterization framework provides a nuanced perspective on the entrainment dynamics captured in the F0 data.

\begin{figure}[t]
\centering
\includegraphics[width=\linewidth]{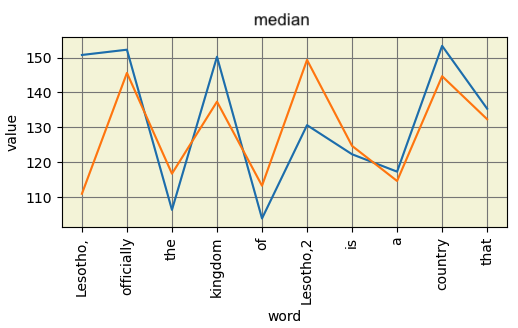}
\caption{F0 contour parameterization (median). On the x-axis is the utterance transcription separated into words and y-axis represents the parameter value. Yellow and blue lines are the plots of F0 parameters of the imitator and model utterance.}
\label{fig:param}
\end{figure}

\subsection{F0 Entrainment}
The quantification of F0 entrainment for each speaker involved assessing the average distance across 40 utterances where the speaker (imitator) mimicked their partner (model speaker). The degree of entrainment, denoted by $E$, is represented by the distance between the parameterized F0 contours of the model and imitated utterances utilising the Dynamic Time Warping (DTW) algorithm with the Euclidean distance metric. A larger $E$ value indicates less entrainment.

The raw value of F0 entrainment $E_{raw}$ is expressed by the distance equation:

\begin{equation}
E_{raw} = \frac{1} {N}\sum_{i=1}^{N}DTW(\mathbf{s}_{i}^{imit}, \mathbf{s}_{i}^{model})
\label{eq1}    
\end{equation}

Here, $N$ denotes the number of utterances imitated by the speaker, $\mathbf{s}_{i}^{imit}$ and $\mathbf{s}_{i}^{model}$ represent the F0 parameter vectors for the utterances uttered by the imitator and the model speaker, respectively. The DTW function gauges the degree of entrainment between these parameterized F0 contours.

In the experiments detailed in Sections \ref{subsec:speaker} and \ref{subsec:dyad}, we employed an optimised entrainment function denoted as:

\begin{equation}
E_{opt} = \frac{1}{N} \sum_{i=1}^{N} \text{Norm}\left(\text{DTW}(\mathbf{s}_{i}^{imit}, \mathbf{s}_{i}^{model})\right)
\label{eq2}    
\end{equation}

Here, the DTW distance is normalised by subtracting the mean of all utterance-level DTW samples and dividing by the sample standard error, as defined by the \text{Norm} function. The sample statistics and $N$, representing the number of utterances per speaker, are computed with outliers removed using the quantile method.

To assess the robustness of the distance algorithms, two additional measures were considered: partner distance and other distance, i.e., the distances of real dyads and surrogate dyads. The partner distance is defined as:

\begin{equation}
\text{partner distance} = E_{raw}^{(t, p)}
\label{eq3}
\end{equation}

\noindent where $E_{raw}^{(t, p)}$ signifies the average raw distance between the target speaker and their partner speaker.

Similarly, the other distance is expressed as:

\begin{equation}
\text{other distance} = \frac{1}{C}\sum_{i}^{C}E_{raw}^{(t, i)}
\label{eq4}
\end{equation}

Here, $E_{raw}^{(t, i)}$ represents the average distance between the target speaker and non-partner speakers, and $C$ denotes the number of combinations of the target speaker with other speakers (i.e., non-partners). Our hypothesis posits that the partner distance will be smaller than the other distance, indicative of the entrainment effect in interactions with the partner.

In Section \ref{subsec:dyad}, we investigated the correlation between the inner-dyad difference in English proficiency and the disparity in the degree of entrainment, as represented by Equation \ref{eq5}:

\begin{equation}
\text{inner-dyad distance} = \text{Norm}(E_{raw}^{A} - E_{raw}^{B}) 
\label{eq5}
\end{equation}

Here, $A$ and $B$ denote the speaking partners. The \text{Norm} function transforms the raw distance by dividing it by the mean of $E_{raw}^{A}$ and $E_{raw}^{B}$. A positive inner-dyad distance indicates that speaker $A$ is less entrained to speaker $B$ in the imitation task, and vice versa. 

In the following experiments, we adopt a significance level of p $<$ 0.05.  Following \cite{levitan2011measuring, weise2019individual} results with p $<$ 0.1 are considered to trend towards significance.

\section{Results}
\subsection{Partner vs non-partner experiment}
\label{subsec:partner}
The comparison between partner and non-partner distances, assessed through a paired t-test across multiple F0 features, reveals statistically significant differences. The negative t-statistics in all cases (See Table ~\ref{tab:ttest_results}) indicate that the mean differences for partner pairs are consistently smaller than those for non-partner pairs, suggesting that the chosen F0 features and the distance algorithms' capability to capture the intricate entrainment effect during speech imitation.
\begin{table}[th]
\caption{T-Test Results Comparing Partner Differences to Other Differences}
\centering
\begin{tabular}{@{}lrrlr@{}}
\toprule[1pt]
\textbf{Feature} & \textbf{t} & \textbf{df} & \textbf{p-value} & \textbf{Sig.} \\ 
\midrule
Median & -4.44 & 57 & 2.1e-5 & * \\
Mean   & -4.43 & 57 & 2.2e-5 & * \\
Range  & -3.67 & 57 & 0.0003 & * \\
Slope  & -4.53 & 57 & 1.5e-5 & * \\
Drop   & -3.42 & 57 & 0.0006 & * \\
\bottomrule[1pt]
\end{tabular}
\label{tab:ttest_results}
\end{table}

\subsection{Individual speaker experiment}
\label{subsec:speaker}
The correlation analysis between individual speakers' degree of F0 entrainment and their spoken English proficiency reveals a nuanced relationship between how closely speakers mimic their partners' speech patterns and their L2 English competence. The F0 entrainment here is quantified as a distance, with higher values indicating less entrainment or mimicry of speech patterns.

\begin{figure}[ht]
\centering
\includegraphics[width=\linewidth]{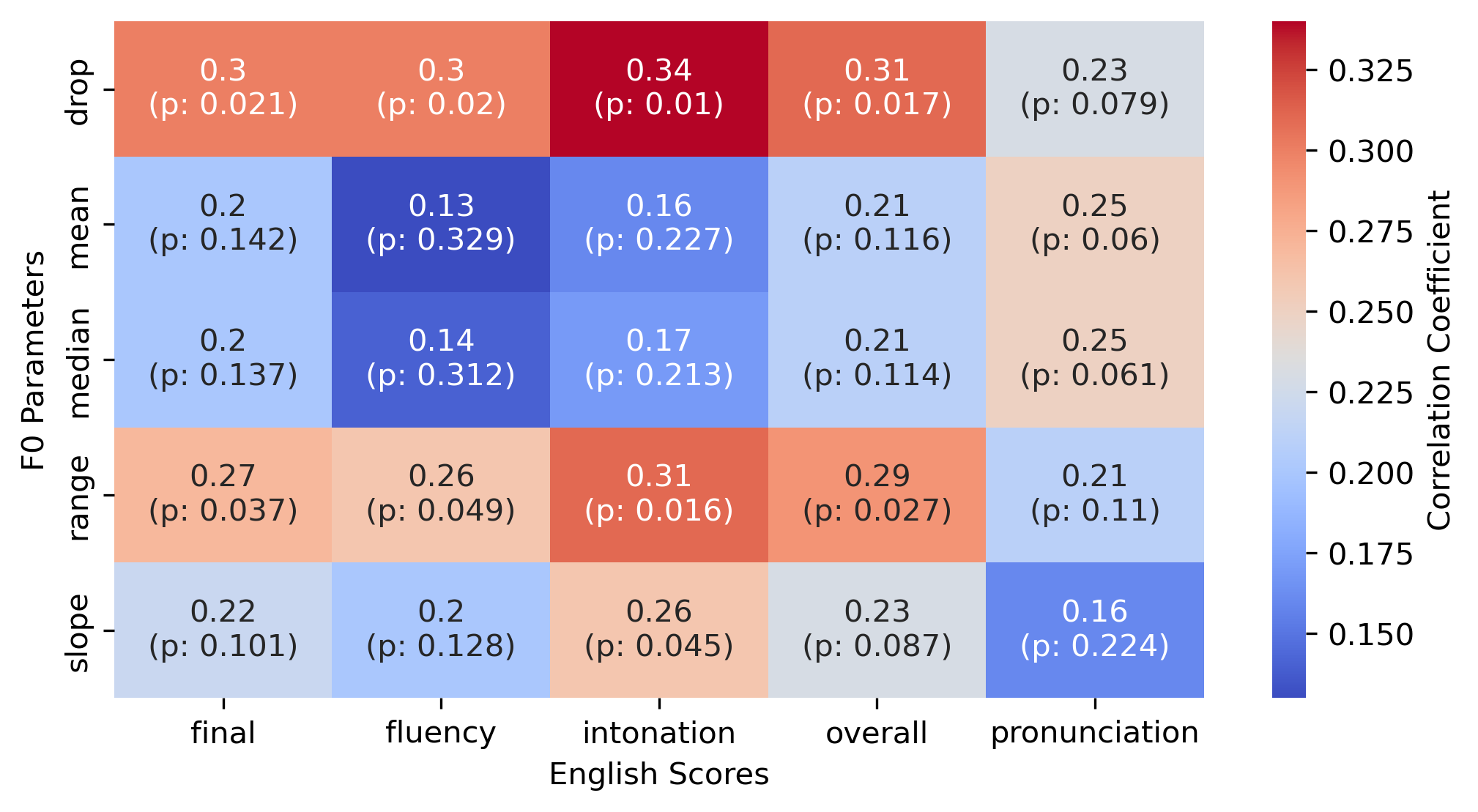}
\caption{Heatmap of Pearson correlation coefficients between F0 entrainment measures and English proficiency scores. Cells indicate correlations of F0 distance (median, mean, range, slope, drop) with language scores (final, fluency, intonation, overall, pronunciation). Colour intensity reflects correlation strength (red: positive, blue: negative), with values in parentheses denoting p-values.}
\label{fig:f0_eng_corr_spk}
\end{figure}

As depicted in Fig. \ref{fig:f0_eng_corr_spk}, "range" and "drop" parameters exhibit the most notable correlations with all language proficiency scores except "pronunciation". "Range" displays moderate associations ($0.26 < r < 0.31$), while "drop" demonstrates stronger correlations ($0.3 < r < 0.34$). "Intonation", among the score parameters, shows statistically significant correlations with "slope" ($r = 0.26$), "range" ($r = 0.31$), and "drop" ($r = 0.34$). Conversely, "median" and "mean" F0 parameters exhibit weaker, less consistent associations with language proficiency, lacking statistical significance, except for a minor connection with "pronunciation".

\subsection{Inner-dyad experiment}
\label{subsec:dyad}

\begin{figure}[ht]
\centering
\includegraphics[width=\linewidth]{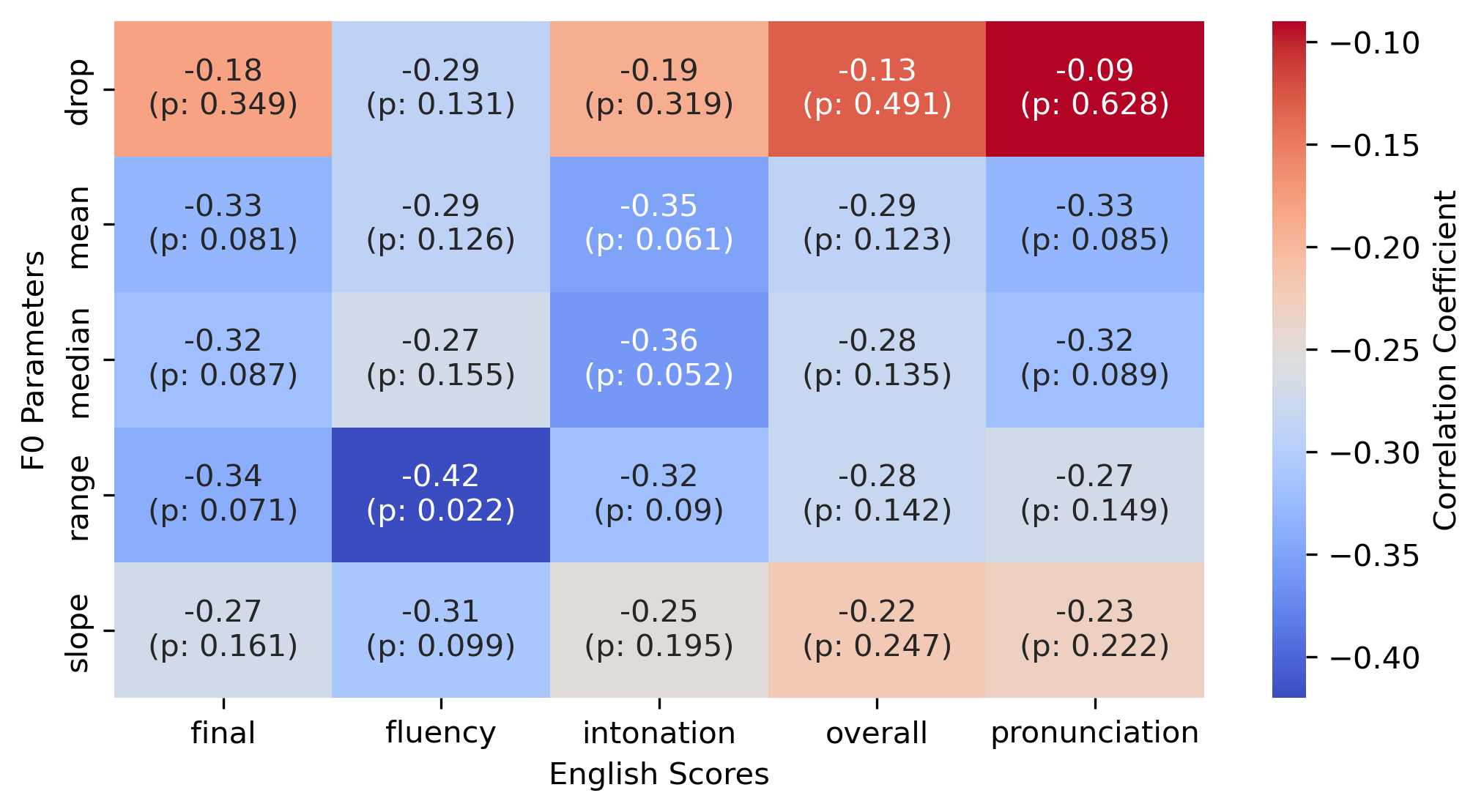}
\caption{Heatmap of Pearson correlation coefficients between inner-dyad differences in English proficiency scores and F0 entrainment measures. Negative correlations are illustrated, with darker shades indicating stronger relationships. The correlation coefficients are accompanied by p-values in parentheses.}
\label{fig:inner_dyad_corr}
\end{figure}

Instead of focusing on absolute English scores and F0 distance values, this experiment explores the intricate relationship between inner-dyad differences in L2 English proficiency scores and corresponding variations in parameterized F0 measures to ascertain whether a more proficient L2 English speaker within a dyad exhibits more or less entrainment with their partner. The findings, presented in Fig.\ref{fig:inner_dyad_corr}, reveal a complex interplay of the two aspects, particularly when considering the collaborative effects within a dyad.

For the "range" parameter and "fluency" score, a significant negative correlation was found (-0.424, p = 0.0219), suggesting that within a dyad, the speaker with higher English fluency tends to be more entrained in terms of pitch variation with their partner. While many other F0 parameters and English score aspects display negative correlations, they do not reach statistical significance, with "final" and "intonation" showing close-to-significant results.







\section{Discussion}

The observation that higher proficiency speakers tend to exhibit less entrainment in terms of pitch variation and declination is a notable finding. It suggests that these individuals have developed a more stable and independent prosodic system in their L2 speech, allowing them to maintain their prosodic identity rather than conforming to the model's patterns. Less proficient speakers have demonstrated relative entrainment flexibility which aligns with \cite{gnevsheva2021phonetic}’s comparison between L2 language speakers and native speakers.

Contrasting with the individual entrainment trends, within dyads, proficient speakers show a tendency for stronger mimicry of pitch range, indicating enhanced entrainment. This could be interpreted as proficient speakers' adaptive linguistic behaviour, showcasing their capability to align more closely with their partner when necessary, perhaps as evidence of stronger sensorimotor adaptation \cite{gambi2013prediction} competence trained through L2 learning.

Our finding also shows that in imitation tasks speakers prioritise dynamic pitch features like range and slope. During imitation, speakers might pay more attention to these prominent and salient prosodic cues as they hold significant emotional and communicative weight, making them more amenable to conscious replication. In contrast, mean and median F0 values, reflecting baseline pitch influenced by physiological characteristics, voice quality, and less linguistic context, might be less subject to conscious manipulation.

The use of DTW distance and F0 parameterization as a measure of entrainment offers a methodological contribution, demonstrating its effectiveness in capturing subtle prosodic alignment phenomena. This quantitative approach could be complemented by qualitative analyses to further understand the subjective aspects of prosodic adaptation including task involvement, partner attractiveness, and motivation \cite{lewandowski2019phonetic}.

The interpretation of the study's findings is subject to certain limitations. Firstly, the relatively small sample size of speakers and language evaluators might exert a disproportionate influence on the results. Moreover, the accuracy of the DTW distance may be compromised by errors in force-alignment. Furthermore, the prosodic parameterization employed was relatively simplistic, relying on a first-order linear fitting that captures only rising or falling intonation patterns, thus overlooking the more complex nuances of prosody.

Future work could explore a wider range of F0 features and apply a Linear Mixed Effects model incorporating variables like sentence length and interaction time. It should also include subjective evaluations of the perceived similarity of the imitated utterances, more evaluators for proficiency ratings, and focus on local entrainment within intra-pausal units to deepen our understanding of L2 speech dynamics.

\section{Conclusions}
The study reveals that higher L2 English proficiency correlates with less F0 entrainment on an individual level but greater mimicry of pitch range within dyadic interaction. This suggests that advanced speakers maintain distinct prosodic patterns while also displaying adaptability in interactive settings. This plasticity sheds light on prosodic dynamics and language acquisition mechanism in L2-L2 interactions. 

\section{Acknowledgements}
\label{sec:tks}

This work was supported by the European Union's Horizon 2020 research and innovation programme under the Marie Skłodowska-Curie grant agreement No 859588. We are grateful to Uwe D. Reichel for generously sharing the pitch parameterization code that served as the foundation for our analysis. We also extend our sincere thanks to Štefan Beňuš, Jana Beňušová, Lucia Mareková, Changyong Min, Qiuwen Zhang, and Fang Liu for their meticulous evaluation of the English language data.

\bibliographystyle{IEEEtran}

\bibliography{mybib.bib}


\end{document}